# Persistent homology-based descriptor for machine-learning potential of amorphous structures

Emi Minamitani[1,2,3*], Ippei Obayashi[3,4], Koji Shimizu[5], Satoshi Watanabe[5]

1. The Institute of Scientific and Industrial Research, Osaka University, Ibaraki 567-0047, Japan
2. Institute for Molecular Science, Okazaki 444-8585, Japan
3. JST, PRESTO, Kawaguchi, Saitama 332-0012, Japan
4. Center for Artificial Intelligence and Mathematical Data Science, Okayama University, Okayama 700-8530, Japan
5. Department of Materials Engineering, The University of Tokyo, Hongo 113-8656, Japan

**Abstract**

High-accuracy prediction of the physical properties of amorphous materials is challenging in condensed-matter physics. A promising method to achieve this is machine-learning potentials, which is an alternative to computationally demanding ab initio calculations. When applying machine-learning potentials, the construction of descriptors to represent atomic configurations is crucial. These descriptors should be invariant to symmetry operations. Handcrafted representations using a smooth overlap of atomic positions and graph neural networks (GNN) are examples of methods used for constructing symmetry-invariant descriptors. In this study, we propose a novel descriptor based on a persistence diagram (PD), a two-dimensional representation of persistent homology (PH). First, we demonstrated that the normalized two-dimensional histogram obtained from PD could predict the average energy per atom of amorphous carbon (aC) at various densities, even when using a simple model. Second, an analysis of the dimensional reduction results of the descriptor spaces revealed that PH can be used to construct descriptors with characteristics similar to those of a latent space in a GNN. These results indicate that PH is a promising method for constructing descriptors suitable for machine-learning potentials without hyperparameter tuning and deep-learning techniques.

# 1. Introduction

High-accuracy prediction of the physical properties of disordered systems such as amorphous materials is significantly challenging in condensed-matter physics. Ab initio calculations, based on quantum mechanics, are versatile and highly accurate. However, the substantial computational cost limits the application of these calculations to realistic amorphous structures. In recent years, machine-learning potentials have emerged as an attractive method to resolve the trade-off between computational cost and accuracy.

Machine-learning potential involves the construction of a surrogate model of the relationship between atomic coordinates and physical properties, which provides an alternative to computationally intensive ab initio calculations. Various machine-learning models have been developed over the past decade, including Gaussian process regression[1,2], artificial neural networks[3–6], and their derivatives[7–12]. A key requirement of machine-learning potential models is the development of a suitable method for converting atomic coordinates into vector data.

Predictions derived from machine-learning potentials should adhere to physical laws. For example, the total energy of a system is invariant to the spatial translation, rotation, reflection, and permutation of chemically equivalent atoms. A direct approach to achieve this is to ensure that the vectorial inputs of the machine-learning potential remain invariant under these symmetry operations. Currently, two methods exist for creating vectorial representations of atomic coordinates, which are commonly known as descriptors. The first method involves using handcrafted feature representations such as the smooth overlap of atomic positions (SOAP)[13] and atom-centered symmetry functions[14]. The second method uses graph neural networks (GNN)[5,6,11,12], where essential structural characteristics are extracted by neural networks during training. Using the former descriptors, even a simple machine-learning model can adequately perform predictions. However, the optimal values of numerous hyperparameters in the descriptor, such as the cutoff, shape, and size of the basis functions, should be determined. The latter method has the advantage of not requiring descriptor hyperparameter tuning, but it requires increased model complexity.

This study proposes a descriptor that differs from the aforementioned approaches, that eliminates the need for hyperparameter tuning while predicts the energies of amorphous materials using a simple machine-learning model. In this study, we focus on persistent homology (PH)[15,16], a recently developed computational topology technique. PH effectively extracts topological and geometrical characteristics and has been applied to structural analyses in wide-ranging fields, from glass to bio-related materials[17–22]. In addition, recent studies have demonstrated that PH effectively identifies correlations between the physical and structural properties of materials[23–29].

Herein, we provide a brief overview of the fundamental concepts of PH. In mathematics, a shape is considered a topological space, $X$, consisting of a set of points (in our case, atoms) and a topology. The topological characteristics of $X$ are represented by holes in space that are encoded in an algebraic structure called the *k*-th homology group $H_k(X)$. A nested sequence of topological space, $X_1 \subseteq X_2 \subseteq \cdots X_n$, results in the series $H_k(X_1), H_k(X_2) \cdots H_k(X_n)$. The concept of PH involves monitoring the elements of $H_k(X_i)$ as *i* increases, where *i* represents the scale and is often referred to as 'time'. The procedure used to obtain this sequence is called *filtration,* and is illustrated in Fig. 1.

During the filtration process for the first homology in a two-dimensional system, circles with a radius $r$ are placed at the respective atoms. Subsequently, $r$ is increased gradually. Consequently, the circles begin to intersect, and then edges are set between the centers of the intersecting circles. The growing network formed by these edges defines the sequence of the topological space, $X_i$. A closed path formed by these edges corresponds to a topological feature called a *cycle*. As the radius continues to increase, the closed path becomes completely covered by circles, which is interpreted as the conversion of the cycle into another type of topological feature called a *boundary*. The features of PH are represented by pairs of birth and death times at which cycles appear and are converted into boundaries. Filtration can be generalized into three or more dimensions using spheres or hyperspheres. A two-dimensional visualization of birth-death time pairs is known as a persistence diagram (PD).

Based on this definition, the PD captures information regarding the bonding state and distribution of atoms, and remains invariant to the spatial translation, rotation, reflection, and permutations of chemically equivalent atoms. Moreover, PD does not require any hyperparameter to be adjusted manually.

In this study, we demonstrate that PH can be used as an input for the machine-learning potential of amorphous structures. The specific target is amorphous carbon (aC). Initially, we demonstrate that the normalized two-dimensional histogram derived from the PD can predict the average energy per atom of aC at various densities, even when using a Ridge model. Moreover, we demonstrate that the prediction accuracy is improved by utilizing a convolutional neural network (CNN) model. Subsequently, we compare the PH descriptors, conventional handcrafted descriptors, and latent space of the GNN-based architectures. SOAP descriptors are selected for the conventional handcrafted descriptor, whereas SchNet is used for the GNN-based architecture. We reveal the differences in how each descriptor captures structural characteristics. The 2D maps obtained from the dimensionality reduction of the PH descriptors resemble those of the SchNet descriptors but differ from those of the SOAP descriptors. Based on these results, we conclude that PH can construct descriptors with characteristics similar to those of a latent space in a GNN, without deep learning and hyperparameter tuning.

## 2. Methods

2.1 Datasets

The aC dataset was generated by ab initio calculations using the VASP software[30–33]. We utilized the LDA exchange-correlation functional and PAW potential for carbon. Melt-quench simulations were performed to generate amorphous and liquid-state structures. A simple cubic lattice with 216 carbon atoms was selected as the initial state. Simulations were conducted at densities of 1.5, 1.7, 2.0, 2.2, 2.4, 2.6, 2.8, 3.0, 3.2, 3.4, and 3.5 g/cm$^3$ to generate different structures. The NVT ensemble[34,35] was used for all the melt-quench simulations, and density was adjusted by varying the size of the simulation cell. A time step of 1 fs was used for the simulations. For all densities, only the Γ points were sampled in k-space. To increase the structural diversity, six independent simulations were performed.

During the melt-quench simulations, the temperature was increased from 300 to 9000 K in 2 ps to melt the carbon. Equilibrium molecular dynamics (MD) was conducted at 9000 K for 3 ps to form the liquid state, followed by a decrease in temperature to 5000 K in 2 ps, and equilibration for 2 ps. Finally, the temperature was decreased from 5000 to 300 K in 2 ps to generate the amorphous structure. During this process, 30 snapshots were obtained from the equilibrium MD trajectory at 9000 K, 100 from the cooling process between 9000 and 5000 K, 25 from the equilibrium MD trajectory at 5000 K, and 100 from the cooling process between 5000 and 300 K, yielding 16,830 data points.

Additionally, the data for diamond structures containing 216 atoms at densities of 2.4, 2.6, 2.8, 3.0, 3.2, 3.4, and 3.5 g/cm$^3$ were prepared. Further data on the diamond structures were obtained from the 80 snapshots of the 2 ps equilibrium MD trajectory at 300 K, yielding 560 data points.

To validate the predictions for larger structures, we generated the data for 512-atom systems using the same procedure as that for the 216-atom systems. A single simulation was conducted for each density. The number of data points was 2,805 for the amorphous and liquid states. These datasets can be downloaded from Zenodo (DOI: 10.5281/zenodo.7905585).

2.2 Persistent homology

In this study, we utilized the HomCloud code[36,37] to analyze the PH of the amorphous models. Herein, we focused on the first homology. To create the input for the machine-learning potential, we converted the PDs into two-dimensional histograms using a 128×128 mesh. The square region of $[0.0, 8.0]_{birth} \times [0.0, 8.0]_{death}$ was targeted. Because the PD represents the number and size of ring structures, the histogram values depend on the system size. Consequently, the histogram was transformed into a probability distribution and z-score standardized to ensure that the descriptors were independent of model size. Because of the small size of the model, periodic boundary conditions were applied when calculating the PDs.

2.3 SOAP descriptors

The SOAP descriptors were calculated using DScribe[38]. The hyperparameters for the SOAP descriptor include the cutoff radius (rcut), number of radial basis functions (nmax), maximum degree of spherical harmonics (lmax), and standard deviation of the Gaussian function used to expand the atomic density (sigma). Optuna[39] was used to search for the optimal hyperparameter values, which were determined to be rcut = 4.1, nmax = 8, lmax = 9, and sigma = 0.16 (see Fig. S1 in the Supplementary Material). These hyperparameters resulted in a SOAP descriptor dimension of 360. The average value across the axis representing the number of atoms was used to evaluate the descriptor space.

2.4 SchNet

SchNet[5] was selected as a representative model of the neural network potential based on GNN. SchNetPack[40] was used to implement the SchNet architecture. SchNet comprises four primary components, as illustrated in Fig. 2 of Ref. 5: Atom embedding, atom-wise layers, interaction blocks, and continuous-filter convolution with filter-generating networks. In SchNet, a set of feature vectors $X^0 = (x_1^0, x_2^0, \ldots x_n^0)$ for the respective atoms was generated using an atom-embedding layer based on the chemical species of the atoms. Here, $n$ denotes the number of atoms. Embeddings was initialized randomly. The feature vector $X^0$ was updated by passing through each component layer. The feature vectors at the $l$th layer are denoted by $X^l = (x_1^l, x_2^l, \ldots x_n^l)$. In addition to these feature vectors, a vector representing the local environment of each atom was constructed within the interaction blocks using filter generation networks based on the distance between the central atom i and neighboring atom j. The $k$th element of the vector is defined as

$$e_k(r_i - r_j) = \exp\left(-\gamma(\|r_i - r_j\| - \mu_k)^2\right),$$

where $\mu_k$ represents the center of the Gaussian function. In this study, $\gamma$ was set to 0.1 Å$^{-2}$ and $\mu_k$ to 60 different values at equal intervals in the range of 0 – 5 Å, yielding a 60-dimensional vector. Subsequently, this vector was passed to a fully connected neural network to produce an output vector, $W^l(r_i - r_j)$. The interaction of atoms incorporated by a continuous-filter convolution is defined as

$$x_i^{l+1} = \sum_{j \in N(i)} x_j^l \circ W^l(r_i - r_j),$$

where $N(i)$ represents the set of neighbors of atoms i.

By updating the learnable weight parameters in the architecture with a combination of the four components during the training procedure, the network generated descriptors that capture the structural characteristics to determine the total energy. In this study, the output from the last interaction block of the trained SchNet model was considered the descriptor generated by SchNet. Because the descriptors were computed for each atom, the resulting descriptors formed an $n \times D$ matrix, where $D$ represents the descriptor dimension which was set to 100. To evaluate the descriptor space, the average value of the $D$-dimensional vector along the axis of the number of atoms was used.

2.5 Ridge Regression

Ridge regression for the total energies, based on the PD-generated descriptors, was conducted using the Scikit-learn package[41]. An intercept term was included in all the regression models. Because all the input values were nonnegative, the regression coefficient was directly interpreted. The regularization parameter was set to 200.0.

2.6 CNN Model

The CNN model was implemented using PyTorch[42]. The model comprises three convolution layers followed by max-pooling layers and two fully connected layers. The ReLU function was used as the activation function in the model. The channel size of the first two convolution layers was 64, whereas that of the last layer was 32. The filter size was 3 × 3 for all the convolution layers.

During the training process, loss function was evaluated based on the mean squared error between the predicted and actual mean energy per atom. The weights and biases in the convolution and fully connected layers were optimized using stochastic gradient descent (SGD) with a Nesterov momentum of 0.9 and a weight decay of 0.001. The milestones were set at 200, 400, 800, and 1200, and the number of epochs was set to 1600. When the epoch reached the milestones, the learning rate reduced by a factor of 0.5. The initial value of learning rate was set to 0.0015.

2.7 Neural Network Potential Based on SOAP Descriptors

Neural network potentials based on SOAP descriptors were implemented using PyTorch. The model consists of two fully connected hidden layers with 40 nodes in each layer. During the training process, loss function was evaluated based on the mean squared error between the predicted and actual mean energy per atom. The weights and biases in the model were optimized using the Adam optimizer. The milestones were set at 200, 500, and 1000, and the number of epochs was set to 2000. When the epoch reached the milestones, the learning rate reduced by a factor of 0.5. The initial value of learning rate was set to 0.001.

3. Results and Discussion

First, we demonstrated that the mean energy per atom in amorphous/liquid carbon can be predicted using PDs. As described in the Methods section, we constructed descriptors based on PH by converting the PDs into two-dimensional histograms. These PH descriptors were used as inputs for the Ridge regression and CNN models. Figures 2 (a) and (b) show the training and test results of the respective models for the 216-atom aC system. In this study, 80% of the data were used as training data, and the remaining 20% were used as test data. Even using a simple Ridge regression model, the mean energy per atom in both amorphous and liquid carbon was predicted by the PH descriptors. The root mean squared error (RMSE) for the test data was 59.2 meV/atom. The prediction accuracy improved by utilizing the CNN model, and the RMSE for the test data decreased to 40.7 meV/atom.

As shown in Fig. S2, both ridge regression and CNN predictions were valid when the diamond structures were included in the dataset. This result highlights that the PH descriptors effectively capture the correlation between the atomic structures and the energies of various structures.

Machine-learning potentials should maintain their prediction accuracy when applied to a system larger than that of the training data. Therefore, we predicted the mean energy per atom in aC for a system with 512 atoms. As shown in Figures 2 (c) and (d), both the Ridge regression and CNN models accurately predicted the energies based on the PH descriptor.

Subsequently, we compared these results with those obtained using other methods. SOAP and SchNet were used as examples of a conventional handcrafted representation and GNN model, respectively. Figure 3 shows the results of the training and test for the 216-atom aC system using these methods. Among the investigated models, the combination of the SOAP descriptor and neural network exhibited the highest accuracy, and the RMSE for the test data was 16.4 meV/atom. The prediction accuracy is sensitive to the selection of hyperparameters. As shown in Fig. S3, different hyperparameters can reduce prediction accuracy. For the SchNet model, the RMSE for the test data was 30.5 meV/atom, which falls between those obtained from the SOAP+NN and PH descriptor+CNN models.

Beyond the comparison of the prediction accuracy, we investigated how these descriptors capture the structural characteristics of amorphous structures using dimensionality reduction. As described in the Methods section, the average values of the descriptors for the atoms were used to evaluate the descriptor space constructed using SOAP and SchNet. We note that this procedure did not cause a significant loss of information. As shown in Fig. S4, the mean energies per atom were accurately predicted by the averaged descriptors using a neural network. Figure 4 shows the t-SNE analysis results. The colors in Figures 4(a)–(c) correspond to density of aC, whereas those in Figures. 4(d)–(f) correspond to the energy of aC. In the t-SNE 2D maps of the SOAP descriptor space, descriptors were grouped into discrete island-like structures based on their densities (Figure 4(a)).

Similarly, the t-SNE 2D maps for SchNet exhibited island-like structures; however, the distribution of each island was broader than that in the case of SOAP, and several islands partially overlapped (Figures 4(b) and (e)). The 2D visualizations of the PH descriptor space exhibited the most continuous distributions with respect to both energy and density, as shown in Figures 4(c) and (f). These characteristics of the 2D visualizations were valid when different hyperparameters values were used in the t-SNE analysis (Figures S5 and Fig. S6). Moreover, principal component analysis (PCA) exhibited similar results. As shown in Figure 5, the 2D visualization of the SOAP descriptor using PCA revealed a discrete distribution, whereas those of the PH and SchNet descriptor spaces were continuous.

Notably, the above-mentioned differences between the SOAP and PH/SchNet descriptors were independent of the hyperparameters used in the SOAP descriptors. As shown in Figure S7, the dimensionality reduction of the SOAP descriptor used in Figure S3 had a discrete distribution similar to those shown in Figures 5 and 6. This indicates that the continuous distribution originated from the additional information captured by the PH and SchNet descriptors but not by the SOAP descriptor. One promising candidate for this type of information is the topological structure, which is intrinsically incorporated into PH and GNN.

Based on these results, we concluded that the extraction of structural characteristics by PH is similar to that by GNN, but different from that by SOAP. Thus, PH can construct a descriptor similar to that of a GNN without deep learning and hyperparameter tuning.

These characteristics of the PH descriptor can be used to reduce the complexity of machine-learning models for predicting the energy of amorphous materials. Another advantage of this descriptor is its prediction interpretability. In the case of Ridge regression, we can estimate the contribution of each birth–death pair from the value of the Ridge coefficient. Figure 6(a) shows the projection of the Ridge coefficient onto the mesh grid used to generate a 2D histogram of the PD. The birth–death pairs in the dark blue region (the large negative coefficient region) corresponded to the local structures that reduce energy. In contrast, those in the dark red region (the large positive coefficient region) corresponded to local structures that increase energy.

By tracing the filtration procedure, we can identify the candidates of the cycle corresponding to each birth–death pair. This enabled the inverse analysis of PH, which was used to determine the atomic arrangements corresponding to the low- and high-energy regions. In some cases, several candidates exhibited equivalent topological characteristics. Various concepts[43,44] exist for selecting a representative cycle for these cases. In this study, the stable volume[44] method was selected, which provides the tightest cycles with robustness against noise.

Figure 6(b) shows the visualization of the high- and low-energy regions obtained by the inverse analysis based on stable volumes. The birth–death pairs in the region where the Ridge coefficient is larger (lower) than 0.003 (-0.0045) were defined as the local structure with high (low) energy. The structure shown in Figure 6(b) was obtained from the equilibrium MD trajectory at 5000 K during the melt-quench simulation, and the mass density of the structure was 2.4 g/cm$^3$. At 5000 K, high-energy liquid- and low-energy amorphous-like structures are expected to mix, which agrees with the results shown in Figure 6(b).

We conducted the same analyses on the remaining 24 samples obtained from the same trajectory. We identified the local structures corresponding to the high- and low-energy regions and evaluated the bond lengths and angles of the respective local structures. Histograms of the bond lengths and angles for the 25 samples are shown in Figures 6(c) and (d). The histograms clearly show differences in the local structures assigned to the low- and high-energy structures. In the bind length histogram, the low-energy structures exhibited a peak at approximately 1.5 Å. In contrast, the peak for the high-energy structures was significantly broader. The histograms of the bond angles for the low- and high-energy structures differed between 50° and 80°. These differences in the histograms indicate that the high-energy structures possess greater randomness in terms of bonds and angles. These results verify the hypothesis that PH effectively extracts the correlation between geometrical structures and their energies.

## 4. Conclusion

In this study, we demonstrated that the descriptors obtained from PH can predict the mean energy per atom in aC using a simple ridge regression model. The prediction accuracy was improved by utilizing a CNN model. Furthermore, we compared the results with those obtained using other methods. As examples of a conventional handcrafted representation and a GNN model, we used SOAP and SchNet, respectively. We demonstrated that the PH descriptor and latent space in the trained SchNet model exhibited similar characteristics. We visualized this point using the dimensionality reduction of the descriptor spaces obtained from PH, SOAP, and SchNet. The t-SNE 2D maps of the SOAP and SchNet descriptor spaces exhibited island-like structures. However, the distribution of each island for SchNet was significantly broader than that for SOAP, and several islands partially overlapped. The 2D visualizations of the PH descriptor space exhibited continuous distributions. A similar result was observed for the 2D maps obtained by PCA. The difference between the SOAP and PH/SchNet descriptor spaces originates from the information on the topological structure, which is intrinsically incorporated into PH and GNN.

Another advantage of the PH descriptor is the interpretability of prediction. Local structures corresponding to the high- and low-energy regions can be determined by utilizing the inverse analysis technique for PH.

In summary, the PH descriptor provides several advantages including a simple model architecture for predicting energy, the absence of hyperparameter tuning in descriptor construction, and interpretability of the prediction. However, several of these aspects require further investigation. First, a method to improve prediction accuracy is required, because the accuracy of prediction by PH is not superior to that of the other methods. One possible reason for this is that PH may lack detailed geometrical information. The complementary use of PH and conventional descriptors may be a promising approach. Another aspect is the construction of PH descriptors for multicomponent systems. The exploration of these possibilities is expected to improve machine-learning potentials based on descriptors that depict both topological and geometrical information.

**Associated content**

Supplementary Material

1. Hyperparameter optimization for SOAP descriptor
2. Prediction results by PH descriptor for the dataset including crystal data
3. Prediction results by SOAP+NN model with hyperparameters different from optimal values
4. Results of prediction using average values of SOAP and SchNet descriptors
5. Results of t-SNE with perplexity =10.0
6. Results of t-SNE with perplexity =50.0
7. Dimensional reduction analyses for SOAP descriptor space with different hyperparameters

**Author information**

Corresponding Author

   Emi Minamitani, The Institute of Scientific and Industrial Research,
   Osaka University, Ibaraki 567-0047, Japan
   Email: eminamitani@sanken.osaka-u.ac.jp
   ORCID: 0000-0002-8003-6526

Authors

   Ippei Obayashi, Center for Artificial Intelligence and Mathematical Data Science,
   Okayama University, Okayama 700-8530, Japan
   Email: i.obayashi@okayama-u.ac.jp
   ORCID: 0000-0002-7207-7280

   Koji Shimizu, Department of Materials Engineering, The University of Tokyo,
   Bunkyo, Tokyo, 113-8656, Japan
   Email: shimizu@cello.t.u-tokyo.ac.jp
   ORCID: 0000-0001-5622-9582

   Satoshi Watanabe, Department of Materials Engineering, The University of Tokyo,
   Bunkyo, Tokyo, 113-8656, Japan
   Email: watanabe@cello.t.u-tokyo.ac.jp
   ORCID: 0000-0002-8069-6938

**Notes**

The authors declare no competing financial interests.

**Acknowledgements**

This study was supported by PRESTO Grants (Grant Numbers JPMJPR1923, and JPMJPR2198), JST, and KAKENHI Grants (Grant Numbers 23H04470, 23H04100, 22H05106, 21H01816, 21H05552, 19H02544, 19H00834, and 20H05884), MEXT, Japan. The calculations were performed using a computer facility at the Research Center for Computational Science (Okazaki, Japan).

Figure captions

Fig.1
Schematic of the filtration procedure used to obtain PD from the data points. In this example, there are five points (a, b, c, d, and e). The PH group obtained by filtration is represented by two birth–death pairs, A and B, in the PD. Values of birth and death time are 0.96 and1.02 for A and 1.05 and 1.30 for B, respectively. Pair A (B) corresponds to the birth and death of the cycle defined by the closed path a-b-c-a (c-b-d-e-c). The gray shaded polygon in the growing sequence of the topological space indicates that the closed path formed by the edges of the polygon is completely covered by circles during the filtration process.

Fig.2
Calculation results and predictions by the machine-learning model based on PH descriptors: Training and test results of the mean energies per atom in aC. The red dots represent the comparison results for the test data, while the blue dots represent those for the training data. Results from the (a) Ridge regression and (b) CNN models are shown, along with (c) and (d) their predictions of the mean energies per atom for larger aC systems, respectively.

Fig. 3
Calculation results and predictions by the machine-learning model based on the SOAP descriptor and SchNet. Training and test results of the mean energies per atom in aC with red (blue) dots representing the comparison results for the test (training) data. Results obtained by (a) the NN based on SOAP descriptors and (b) the SchNet model.

Fig. 4
Dimensionality reduction results obtained by t-SNE for the SOAP, PH, and SchNet descriptors. The perplexity in the t-SNE model is 30.0. In panels (a)–(c), the color of the data points corresponds to the density values, whereas in panels (d)–(f), it corresponds to the energy values.

Fig. 5
Dimensionality reduction results obtained by PCA for the SOAP, PH, and SchNet descriptors. In panels (a)–(c), the color of the data points corresponds to the density values, whereas in panels (d)–(f), it corresponds to the energy values.

Fig. 6

Inverse analysis results based on the PH descriptor: (a) Projection of the Ridge coefficient onto the mesh grid used to generate the 2D histogram of the PD. (b) Visualization of the high- and low-energy regions in a 512-atom aC sample determined by inverse analysis. (c) and (d) Histograms of the bond lengths and angles in the high- and low-energy regions determined by inverse analysis. Data from 25 samples are aggregated and visualized as histograms.

Figures

Fig.1

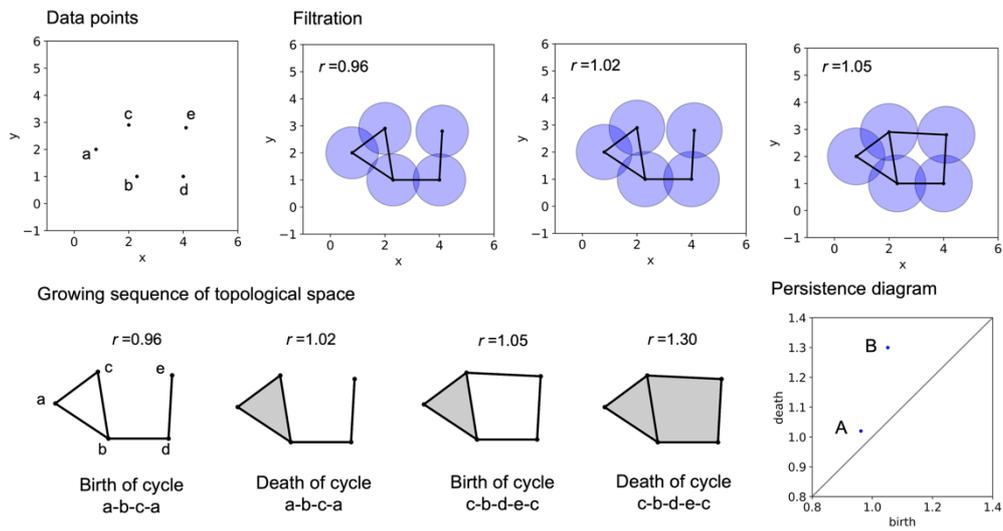

Fig.2

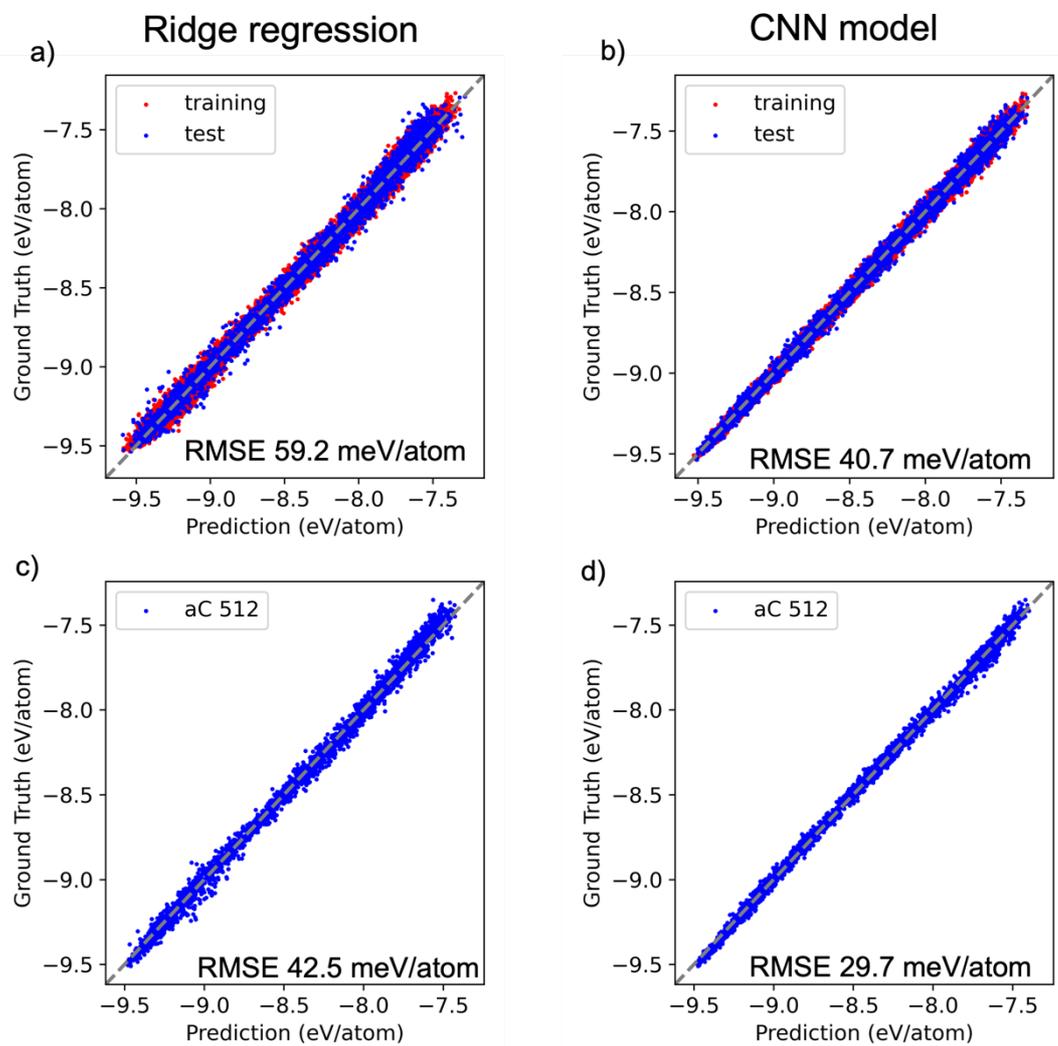

Fig. 3

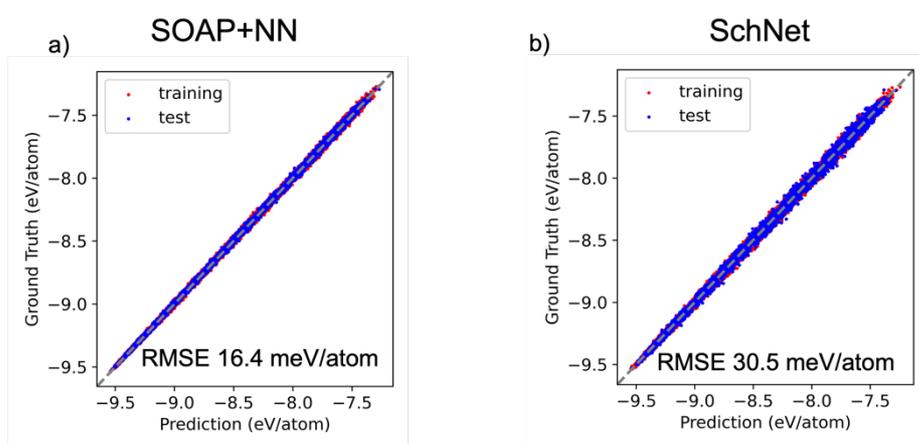

Fig. 4

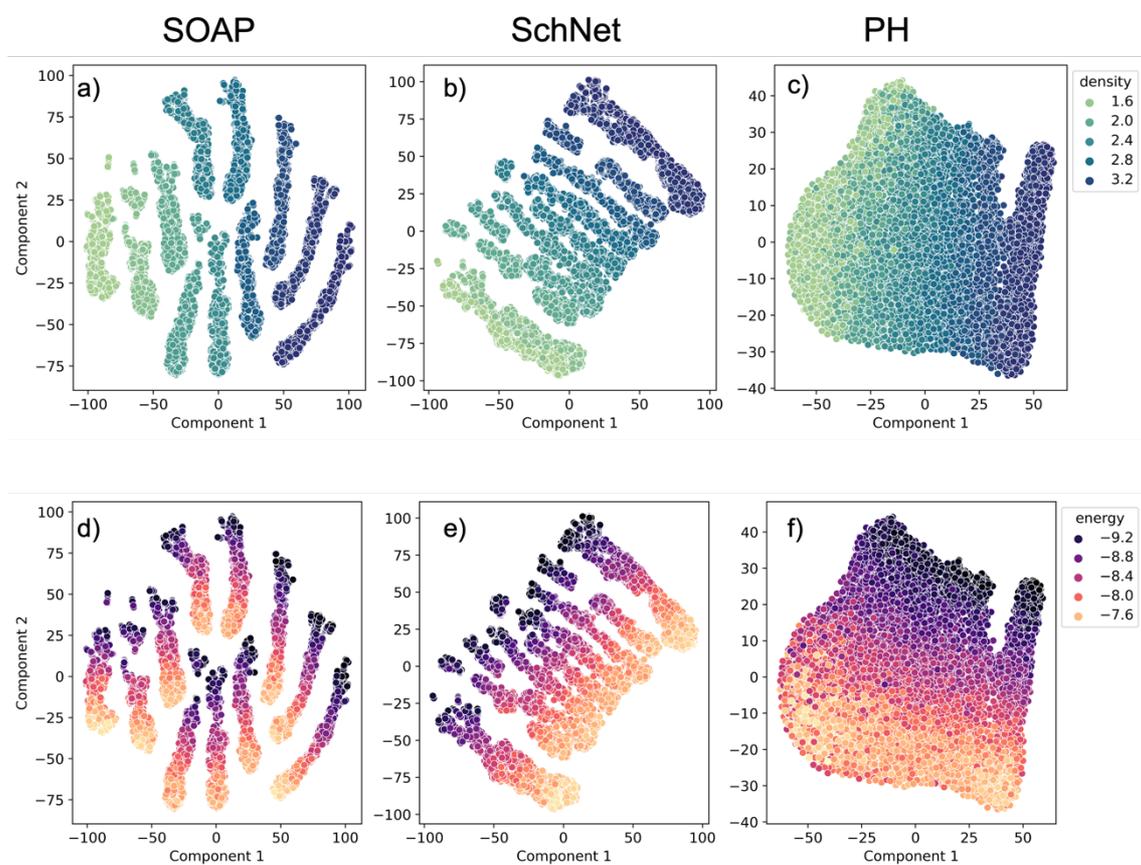

Fig. 5

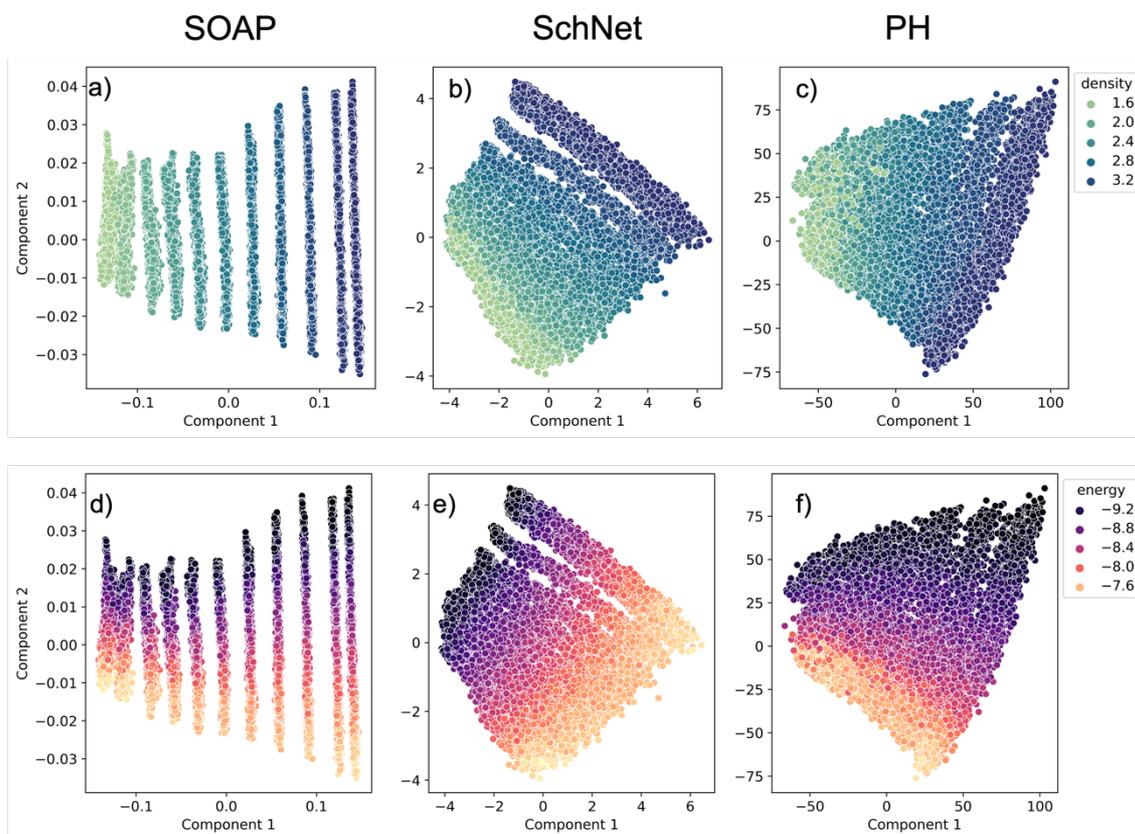

Fig. 6

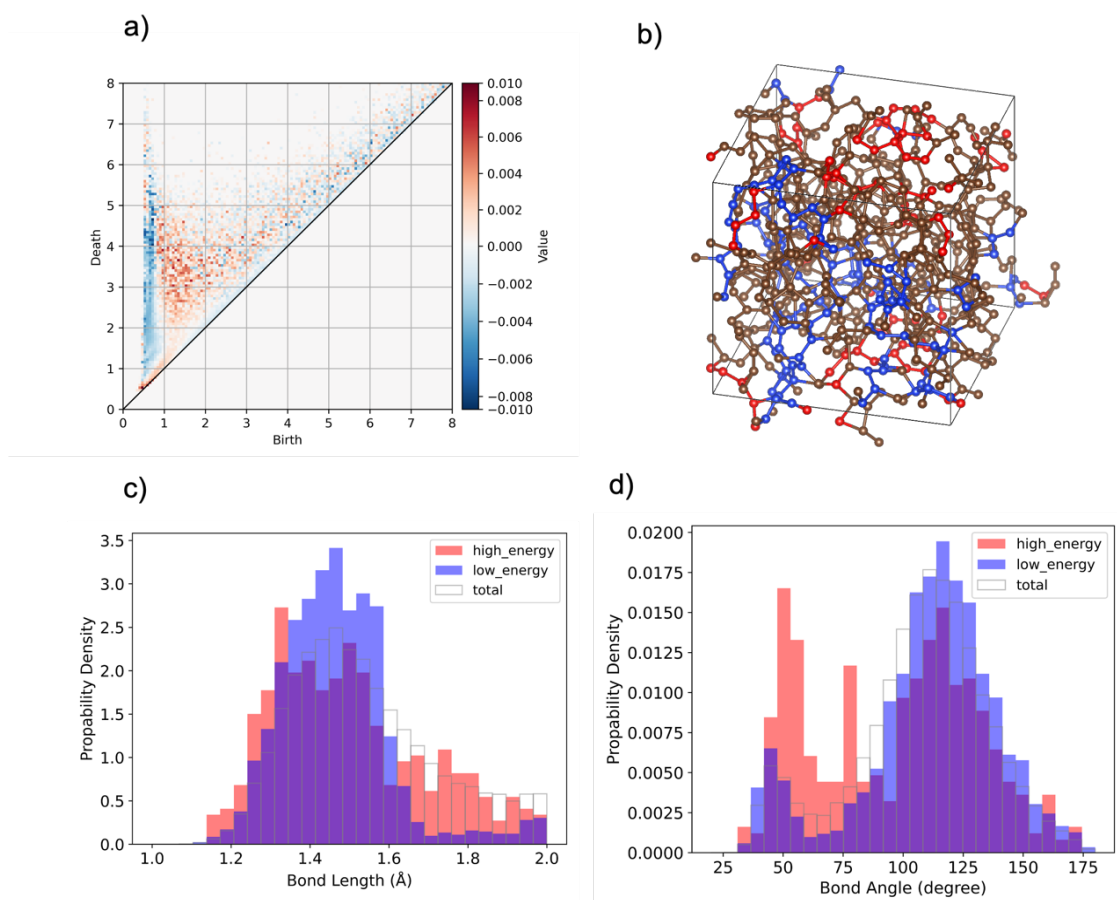